\documentclass[pdflatex,sn-mathphys]{sn-jnl}

\usepackage{subfig}
\usepackage{amsfonts}
\usepackage[bbgreekl]{mathbbol}
\DeclareSymbolFontAlphabet{\mathbbm}{bbold}
\DeclareSymbolFontAlphabet{\mathbb}{AMSb}%

\jyear{2021}%

\theoremstyle{thmstyleone}%
\theoremstyle{thmstyletwo}%

\theoremstyle{thmstylethree}%

\begin{document}

\title[Non-rigid Point Cloud Registration for Middle Ear Diagnostics]{Non-rigid Point Cloud Registration for Middle Ear Diagnostics with Endoscopic Optical Coherence Tomography}


\author*[1,3]{\fnm{Peng} \sur{Liu}}\email{peng.liu@nct-dresden.de}
\author[2,3]{\fnm{Jonas} \sur{Golde}}
\author[3,4]{\fnm{Joseph} \sur{Morgenstern}}
\author[1]{\fnm{Sebastian} \sur{Bodenstedt}}
\author[1]{\fnm{Chenpan} \sur{Li}}
\author[1]{\fnm{Yujia} \sur{Hu}}
\author[5]{\fnm{Zhaoyu} \sur{Chen}}
\author[1,2]{\fnm{Edmund} \sur{Koch}}
\author[1,4]{\fnm{Marcus} \sur{Neudert}}
\author[1,3]{\fnm{Stefanie} \sur{Speidel}}

\affil*[1]{\orgdiv{Translational Surgical Oncology}, \orgname{National Center for Tumor Diseases}, \orgaddress{ \city{Dresden}, \postcode{01307}, \country{Germany}}}

\affil[2]{\orgdiv{Clinical Sensoring and Monitoring}, \orgname{TU Dresden}, \orgaddress{\city{Dresden}, \postcode{01307}, \country{Germany}}}


\affil[3]{\orgdiv{Else Kröner Fresenius Center}, \orgname{TU Dresden}, \orgaddress{\city{Dresden}, \postcode{01307}, \country{Germany}}}

\affil[4]{\orgdiv{Ear Research Center Dresden}, \orgname{TU Dresden}, \orgaddress{\city{Dresden}, \postcode{01307}, \country{Germany}}}




\abstract{
\textbf{Purpose:} 
Middle ear infection is the most prevalent inflammatory disease, especially among the pediatric population. Current diagnostic methods are subjective and depend on visual cues from an otoscope, which is limited for otologists to identify pathology. To address this shortcoming, endoscopic optical coherence tomography (OCT) provides both morphological and functional in-vivo measurements of the middle ear. However, due to the shadow of prior structures, interpretation of OCT images is challenging and time-consuming. To facilitate fast diagnosis and measurement, improvement in the readability of OCT data is achieved by merging morphological knowledge from ex-vivo middle ear models with OCT volumetric data, so that OCT applications can be further promoted in daily clinical settings.

\textbf{Methods:} We propose C2P-Net: a two-staged non-rigid registration pipeline for complete to partial point clouds, which are sampled from ex-vivo and in-vivo OCT models, respectively.
To overcome the lack of labeled training data, a fast and effective generation pipeline in Blender3D is designed to simulate middle ear shapes and extract in-vivo noisy and partial point clouds.

\textbf{Results:} We evaluate the performance of C2P-Net through experiments on both synthetic and real OCT datasets. The results demonstrate that C2P-Net is generalized to unseen middle ear point clouds and capable of handling realistic noise and incompleteness in synthetic and real OCT data. 

\textbf{Conclusion:} In this work, we aim to enable diagnosis of middle ear structures with the assistance of OCT images. We propose C2P-Net: a two-staged non-rigid registration pipeline for point clouds to support the interpretation of in-vivo noisy and partial OCT images for the first time. Code is available at: https://gitlab.com/nct\_tso\_public/c2p-net.
}




\maketitle

\section{Introduction}\label{sec1}
\begin{figure}[ht]
    \includegraphics[trim=0cm 5.7cm 6cm 0cm,clip, width=\textwidth]{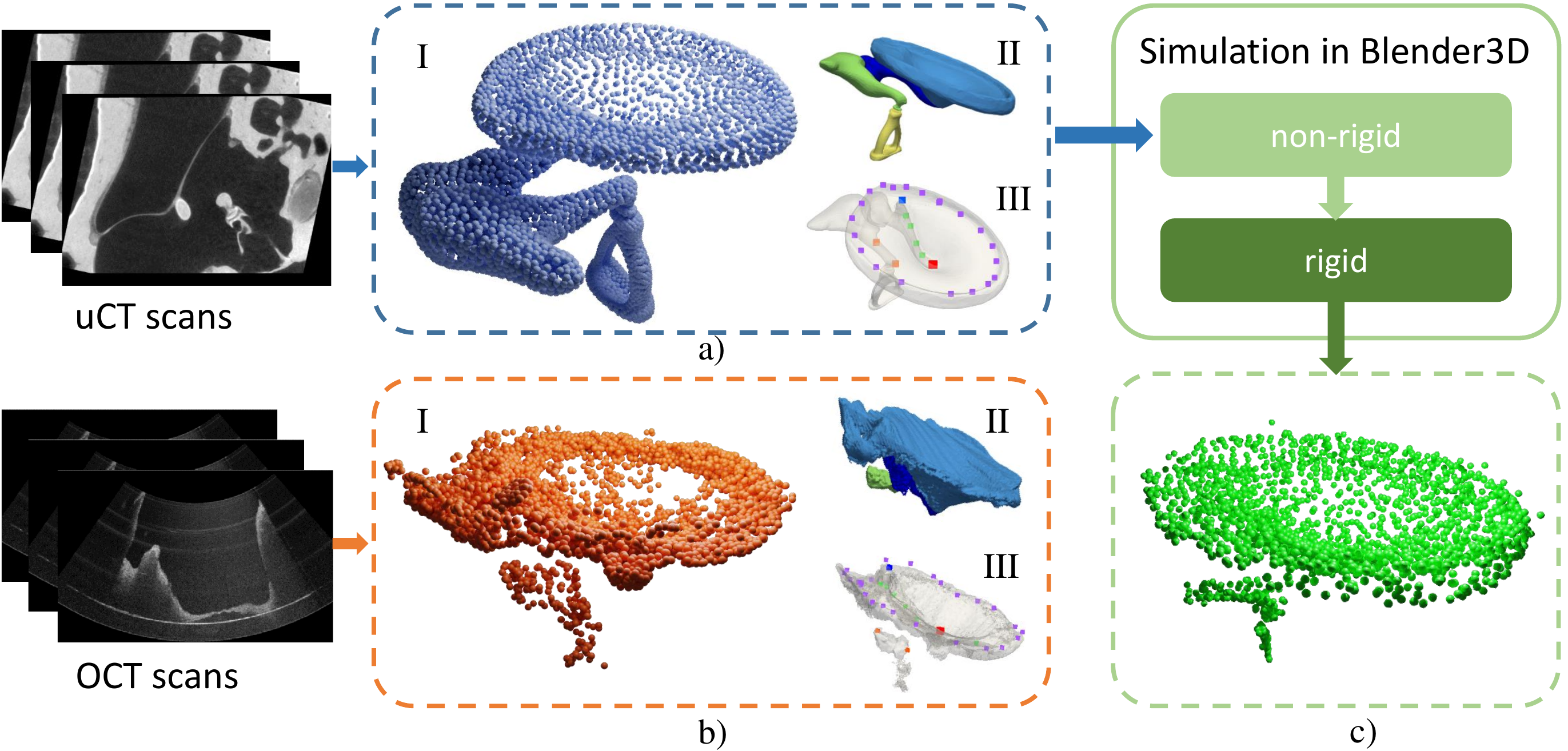}
    \caption{Overview of data samples. In a), I is the ex-vivo template point cloud reconstructed from $\mu$CT scans on the left. II is the segmented middle ear model, each color stands for one structure. III shows the model with sparse labels. Accordingly, I, II and III in b) present in-vivo OCT point cloud, segmented OCT model and labels. The simulation pipeline on the top left takes a complete ex-vivo ear point cloud, performs non-rigid and rigid simulation, and produces synthetic partial and noisy in-vivo point cloud shown in c). }
    \label{data_pipeline}
\end{figure}

The middle ear consists of the tympanic membrane (TM) and an air-filled chamber containing the ossicle chain (OC) that connects the TM to the inner ear. From a functional perspective, the middle ear matches the impedance from air to the fluid-filled inner ear \cite{zwislocki1982normal}.
Middle ear disorders contain deformation, discontinuation of TM and OC, effusion, and cholesteatoma in the middle ear. These may occur because of middle ear infection (e.g.  acute otitis media (AOM), chronic otitis media (COM), otitis media with effusion (OME)) and trauma \cite{won2021c, won2021d}. Most commonly, serious and chronic middle ear disorder may lead to conductive hearing loss and inner ear disorder \cite{snow2009ballenger}. Current middle ear diagnostics methods or tools cover only aspects of the pathology and cannot determine the origin and site of transmission loss, e.g. otoscopy, audiometry, tympanometry, etc. 
As an innovative image technology, endoscopic OCT enables the examination of both the morphology and function of the middle ear in-vivo by the non-invasive acquisition of depth-resolved and high-resolution images \cite{Morgenstern2020, Monroy2019, Monroy2015}. Nevertheless, multiple sources of noise, e.g. the shadow of preceding structures, reduce the quality of the target structures further away from the endoscopic probe, e.g. ossicles. Therefore, the reconstructed 3D models from the OCT volumetric data are usually noisy and difficult to interpret, especially the identification of the deeper middle ear structures such as incus and stapes (see Fig.~\ref{data_pipeline}).

Our aim is to improve the interpretation of the OCT volumetric data by merging the ex-vivo middle ear model reconstructed from micro-Computed Tomography ($\mu$CT) scan of isolated temporal bones and the in-vivo OCT model. 
Note that a single $\mu$CT model was used as a template and fitted to all patient-specific OCT scans.
We first convert all ex- and in-vivo middle ear data to point cloud representations for ease of flexible manipulation of middle ear shapes. Nevertheless, finding one-to-one correspondence between such a point cloud pair is still challenging due to the noise and incompleteness of the OCT model, and the difference between the patient data and the template $\mu$CT data. To tackle these issues, we first employ a neural network that searches the sparse correspondences for the source and target points, then a pyramid deformation algorithm to fit the points in a non-rigid fashion based on the predicted correspondences. The neural network is trained on randomly generated synthetic shape variants of the middle ear that contain random noise and only partial points. This enables the generalization of the neural network to new patient data as well as the adaption to noise and outliers
The main contributions of this work are listed below:
\begin{enumerate}
   \item A generation pipeline in Blender3D that simulates synthetic shape variants from a complete ex-vivo middle ear model, and simulates noisy and partial point clouds comparable to in-vivo data.
   \item C2P-Net: a two-staged non-rigid registration pipeline for point clouds. It demonstrates that C2P-Net registers the complete point cloud template to partial point clouds of the middle ear effectively and robustly.
\end{enumerate}

\section{Related Work}\label{related_work}
In this section, we review methods that analyze 3D point clouds and solve registration problems. 
Learning the geometry of 3D point clouds is the foundation of diverse 3D applications, but is also a challenging task due to the irregularity and asymmetry of point clouds. Recently, approaches that focus on learning point features with neural networks have been widely investigated to overcome such issues. One category of methods is to project the point cloud to a regular representation, e.g. voxel grids~\cite{brock2016generative}, 2D images from multi-views~\cite{su2015multi, feng2018gvcnn}, or combined~\cite{qi2016volumetric} that 2/3D convolutional operations can be performed on top of these intermediate data in Euclidean space. Apart from these, PointNet~\cite{qi2017pointnet} leverages multilayer perceptron (MLP) to obtain pointwise features and a max-pool to aggregate global features but without local information. This drawback is alleviated by PointNet++~\cite{qi2017pointnet++}, which uses PointNet to aggregate local features in a multi-scale structure. In addition, various explicit convolution kernels are applied directly on the point clouds to extract encodings~\cite{thomas2019kpconv, xu2018spidercnn, choy2019fully} in which point features are extracted based on the kernel weights.


Conventional methods solve the point cloud registration task as an optimization problem of transformation parameters. Iterative Closest Point (ICP)~\cite{besl1992method} iteratively calculates a rigid transformation matrix based on updated correspondence set from a last iteration, Non-rigid Iterative Closest Point (NICP)\cite{amberg2007optimal} achieves non-rigid registration by optimizing an energy function including local regularisation and stiffness. Coherent Point Drift (CPD) maximizes the probability of Gaussian Mixture Model (GMM) by moving coherent GMM centroids from the two point clouds. Recently, deep learning-based methods using the aforementioned point cloud encoding techniques have been widely investigated\cite{aoki2019pointnetlk,wang2019non,yew2022regtr,zhu2022neighborhood,li2022non}. 
PointNetLK\cite{aoki2019pointnetlk} extracts semantic point cloud features using MLP, then solves rigid alignment using the Lucas and Kanade algorithm\cite{lucas1981iterative}. PR-Net\cite{wang2019non} follows a similar way of feature extraction while it maps the feature to regular grids and defines the registration task as a correlation problem. RegTR\cite{yew2022regtr} and NgeNet\cite{ zhu2022neighborhood} both introduce attention mechanisms to enable the communication of the extracted point cloud features from the source and target and estimate a global transformation from the predicted correspondences with MLPs. In contrast to this direct way, NDP\cite{li2022non} decomposes the global transformation into sub-motions which are iteratively refined using MLPs at each pyramid level. Inspired by this recent work regarding point cloud analysis and registration, we adopted NgeNet\cite{zhu2022neighborhood} and NDP\cite{li2022non} as the main components to construct our registration pipeline.




%




\section{Methods}
The non-rigid registration of the ex- and in-vivo middle ear point clouds (see Fig.~\ref{data_pipeline}) using neural networks is realized in a supervised learning fashion. Due to lack of real OCT data with ground truth, and the difficulties of finding one-to-one correspondences between the general and ex-vivo model and the noisy and partial in-vivo model, we generated synthetic in-vivo middle ear point clouds based on the ex-vivo $\mu$CT model using a two-step simulation pipeline. For point cloud registration, we propose C2P-Net, which is a two-stage registration method. We first train a neural network on the synthetic samples to explore the correspondence along with an initial rigid transformation matrix, then based on this, a hierarchical algorithm is performed to estimate the non-rigid deformation.

\subsection{Synthetic Data Generation}\label{method:syn_gen}
We start with a complete middle ear model $M_{\mu{CT}}$ reconstructed from $\mu$CT volume as basis for the simulation in Blender3D. 
The non-rigid simulation is performed with the assistance of lattice modifiers attached to each structure with reasonable resolution. The lattice vertices are assigned to various groups according to length, thickness, and width of the global and local shape of a structure. By transforming and rotating various vertice groups with random parameters with boundary conditions, a random shape of a structure can be generated, e.g. a relatively shorter malleus with a longer lateral process. 
The non-rigid simulation step is formulated as $\tilde{T}_{nr}=\mathrm{S}_{nr}(T, p_{nr})$, where $T$ is the template middle ear, $p_{nr}$ are the input parameters, and $\Tilde{T}_{nr}$ is the simulated non-rigid shape variants of $T$.
Next, armature modifiers are attached to each structure and connected in an end-to-end fashion at the articulations. Then, rigid simulation is accomplished by transforming and rotating the armature bones. We denote the rigid simulation as $S_r$. Thus, a random shape variant of the middle ear $\Tilde{T}$ can be obtained which is considered as the in-vivo ear model of a new patient:  $\Tilde{T}=\mathrm{S}_{r}(\Tilde{T}_{nr}, p_{r})$.


In practice, the in-vivo OCT model is usually noisy and only partially visible, which limits the registration methods in finding accurate and robust correspondences. To simulate the real data, we extract random partial patches for each structure of the shape variant $\Tilde{T}$ by a combined probability, which is calculated as the distance of each vertex to the support point of each structure and random Gaussian noise. Additionally, for the sake of simulating incomplete posterior structure, e.g. stapes, caused by the occlusion and shadow from the anterior structures, e.g. ear canal wall,
we determine the number of points of the posterior structures by the distance to the external ear and a random factor. Furthermore, uniform random displacements are applied to the vertices to generate the final noisy and partial in-vivo point cloud $P_{inv}$ (see Fig.~\ref{data_pipeline}).


\begin{figure}[ht]
    \includegraphics[trim=0cm 12.3cm 0.5cm 2.2cm, clip, width=\textwidth]{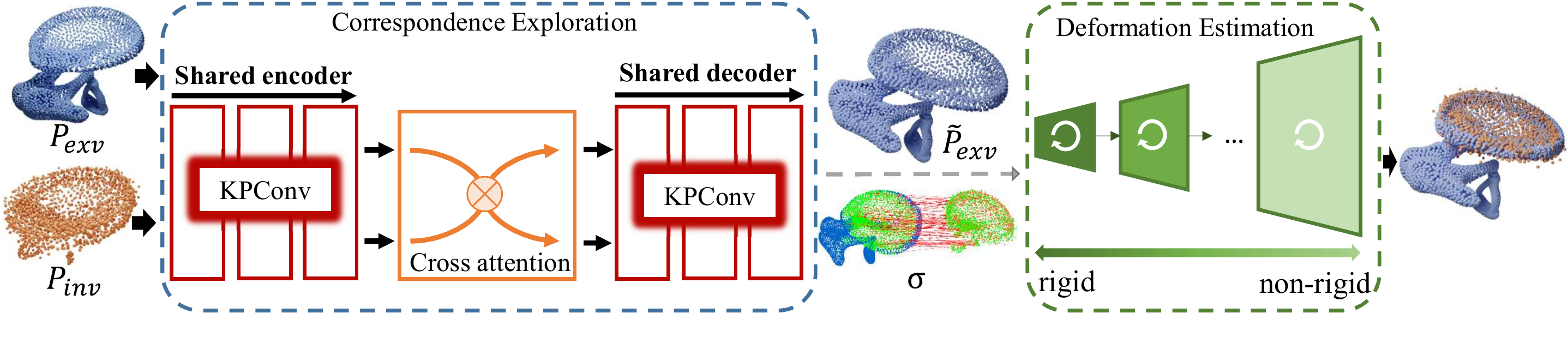}
    \caption{
    Schematic architecture of C2P-Net. The input of the pipeline is the ex-vivo (blue) and in-vivo (orange) point cloud. To explore the correspondence between the two inputs, NgeNet is trained on synthetic samples to extract the features for each point at different levels. These are integrated by a voting algorithm. A feature matching routine is performed to obtain the sparse correspondence set as well as a rigid transformation matrix. On top of the previous outputs, NDP maps the input point clouds to sinusoidal space and decomposes the global non-rigid deformation at each pyramid level into sub-motions. With the increase of sinusoidal frequency, the non-rigid deformation of each point is continuously refined at each level. The registered point cloud is shown on the right side.}
    \label{nn_structure}
\end{figure}

\subsection{C2P-Net}
We delineate the architecture of C2P-Net in Fig.~\ref{nn_structure}. It consists of two components dedicated to two stages: initial rigid registration and pyramid non-rigid registration.
Given an ex-vivo point cloud as a template extracted from a $\mu$CT model: $P_{exv}={\{x_i\in{\mathcal{R}^{3}}\}_{i=1,2,...,N}}$, and a partial point cloud of the simulated in-vivo shape variant: $P_{inv}={\{y_j\in{\mathcal{R}^{3}}\}_{j=1,2,...,M}}$,
we adopted the Neighborhood-aware Geometric Encoding Network (NgeNet)\cite{zhu2022neighborhood} to solve the initial rigid registration task. This stage is formulated as:

\begin{equation}
    \tau,\sigma=\mathrm{NgeNet}(P_{exv}, P_{inv})\quad where\ (u, v) \in\sigma
  \label{methods:ngenet}
\end{equation}
where $\tau\in{SE(3)} $ is the rigid transformation matrix which aligns $P_{exv}$ with $P_{inv}$, and $(u, v) \in\sigma$ is the sparse correspondence set where $u$ and $v$ are indices for points in $P_{exv}$ and $P_{inv}$. Due to the multi-scale structure and a voting mechanism integrating features from different resolutions, NgeNet handles noise well and predicts correspondence robustly.








Based on the previous predicted correspondence set and the rigidly aligned source and target point clouds, we employ the Neural Deformation Pyramid (NDP)\cite{li2022non} to predict the non-rigid deformation of the given point cloud pair. NDP defines the non-rigid registration problem as a hierarchical motion decomposition problem. At each pyramid level, the input points from last level are mapped to sinusoidal encodings with different frequencies: $\Gamma(p^{k-1})=(\mathrm{sin}(2^{k+k_0}p^{k-1}), \mathrm{cos}(2^{k+k_0}p^{k-1}))$, $k$ is the current level number, $k_0$ controls the initial frequency and $p^{k-1}$ is an output point from the last level. 
Lower frequencies at shallower levels represent rigid sub-motion, while higher frequencies at deep levels emphasize non-rigid deformations. In this way, a sequence of sub-motions is estimated from rigid to non-rigid, and the final displacement field is the combination of such a sequence. Formally, we denote the stage as:

\begin{equation}
    \phi_{est}=\mathrm{NDP}_{n,m}(\tilde{P}_{exv}, P_{inv}, \sigma)
  \label{methods:ndp}
\end{equation}
where $\tilde{P}_{exv}$ is the source point cloud ${P}_{exv}$ transformed by $\tau$, $n$ is the number of pyramid layers of the NDP neural network, $m$ is the maximal iteration within a single pyramid layer, and $\phi_{est}$ is the predicted displacement field describing how each point should move to the target. Combined losses are calculated at each iteration, including correspondence loss and regularization loss, and back-propagated to update the weights of each MLP. Of which, the correspondence loss $L_{CD}$ is defined as the Chamfer distance \eqref{chamfer_dist} between $\tilde P_{exv}$ which is masked by the correspondence $\sigma$ and $P_{inv}$.


\begin{equation}
    \mathrm{CD}(A,B) =\frac{1}{\lvert{A}\rvert}\sum_{x_i\in{A}}\min_{y_j\in{B}}\lvert{x_i-y_j}\rvert+\frac{1}{\lvert{B}\rvert}\sum_{y_j\in{B}}\min_{x_i\in{A}}\lvert{x_i-y_j}\rvert
    \label{chamfer_dist}
\end{equation}
\begin{equation}
    L_\mathrm{CD} =\mathrm{CD}(\tilde P_{exv}^{\sigma}, P_{inv})
    \label{loss_cd}
\end{equation}
$\tilde P_{exv}^{\sigma}=\{\tilde P_{exv}[u]\lvert\exists{v}:(u,v)\in\sigma\}$ are the masked ex-vivo points that have correspondence in the target in-vivo point cloud.




\section{Evaluation}


C2P-Net predicts a displacement field to fit the ex-vivo point cloud template to the target in-vivo point cloud which is noisy and partial. At training time, C2P-Net employs an Adam optimizer and ExpLR scheduler and uses supervised learning to learn on a synthetic dataset with 20,000 in-vivo shape variants of the middle ear. During inference, it takes around 3.5s for C2P-Net to react to a new in-vivo shape on RTX 2070 Super. Since the neural network is trained on synthetic data, it is crucial to investigate the generalization of the neural networks on unseen samples as well as the performance on real OCT data.
Thus, we conduct two experiments on synthetic data and real OCT data with various metrics: mean displacement error \eqref{mde}, Chamfer distance \eqref{chamfer_dist}, and landmarks error \eqref{landmarks_error}. Furthermore, we compare our results with other popular non-rigid registration methods, e.g. Non-rigid ICP (NICP) and Coherent Point Drift (CPD).

\begin{equation}
    M_\mathrm{MDE}=\frac{1}{N}\sum_{i}^{N}\lVert{\phi_{est}-\phi_{gt}}\rVert 
    \label{mde}
\end{equation}
$N$ is the size of the test dataset, $\phi_{est}$ is the estimated displacement field of a sample, and $\phi_{gt}$ is the corresponding ground truth.

\begin{table}[h]
\begin{center}
\begin{minipage}{\textwidth}
\caption{Experiments results on synthetic data (SYN) and real OCT data (REAL), including mean displacement error ($M_\mathrm{MDE}$)[mm], landmarks error ($M_\mathrm{L}$)[mm] and Chamfer distance ($M_\mathrm{CD}$)[mm]}
\begin{tabular*}{\textwidth}{@{\extracolsep{\fill}}lccccc@{\extracolsep{\fill}}}
\toprule%
& \multicolumn{3}{@{}c@{}}{SYN} & \multicolumn{2}{@{}c@{}}{REAL}  \\\cmidrule{2-4}\cmidrule{5-6}  %
 & $M_\mathrm{MDE}\downarrow$ & $M_\mathrm{CD}\dagger$ & $M_\mathrm{L}\downarrow$ & $M_\mathrm{L}\downarrow$ & $M_\mathrm{CD}\dagger$   \\
\midrule
ICP$^\ast$  &  2.01     &  1.85   & 1.15  & 4.74     &  3.03 \\
NICP        &  4.61    &  \textbf{0.22}   & 0.87  &  4.39     & 0.68   \\
CPD         &  8.79     & 1.07    & 0.96  &  2.17     &  \textbf{0.059} \\
C2P-Net (ours) &  \textbf{0.781}  & 0.582    & \textbf{0.424}  &  \textbf{0.808}     &  2.43 \\
\botrule
\label{evaluation:stats_table}
\end{tabular*}
\footnotetext[\ast]{rigid registration method(s).}
\footnotetext[\dagger]{baseline methods tend to fall into local minima and ignore the geometry, though the CD is lower.}
\end{minipage}
\end{center}
\end{table}

\begin{figure}[!ht]

    \subfloat[\label{mde:visible_ratio}]{
    \includegraphics[trim=4.6cm 9cm 4.5cm 9cm,clip, width=.492\textwidth]{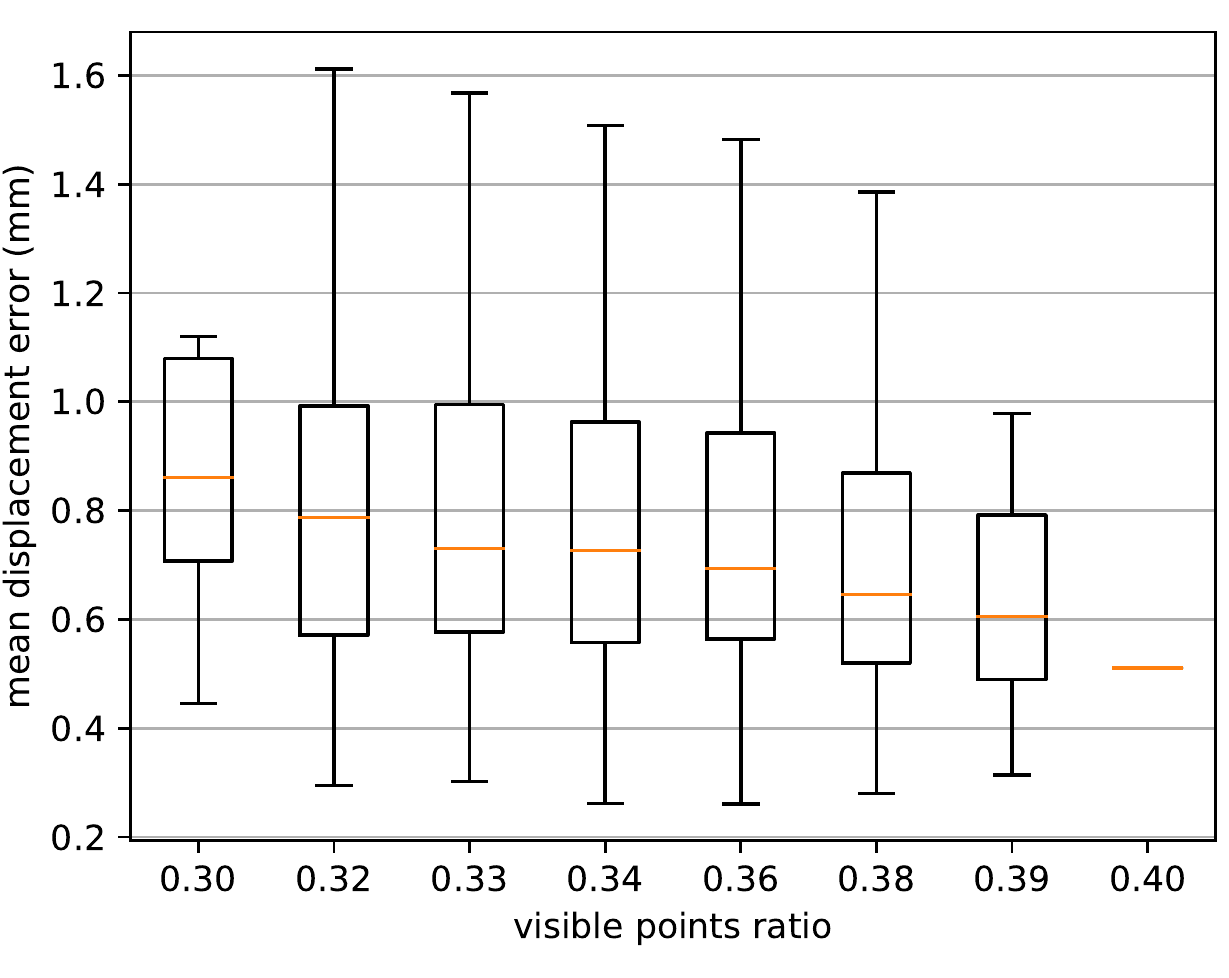}
    }
    \hfill
    \subfloat[\label{mde:initial_registration}]{
    \includegraphics[trim=4.6cm 9cm 4.5cm 9cm,clip, width=.490\textwidth]{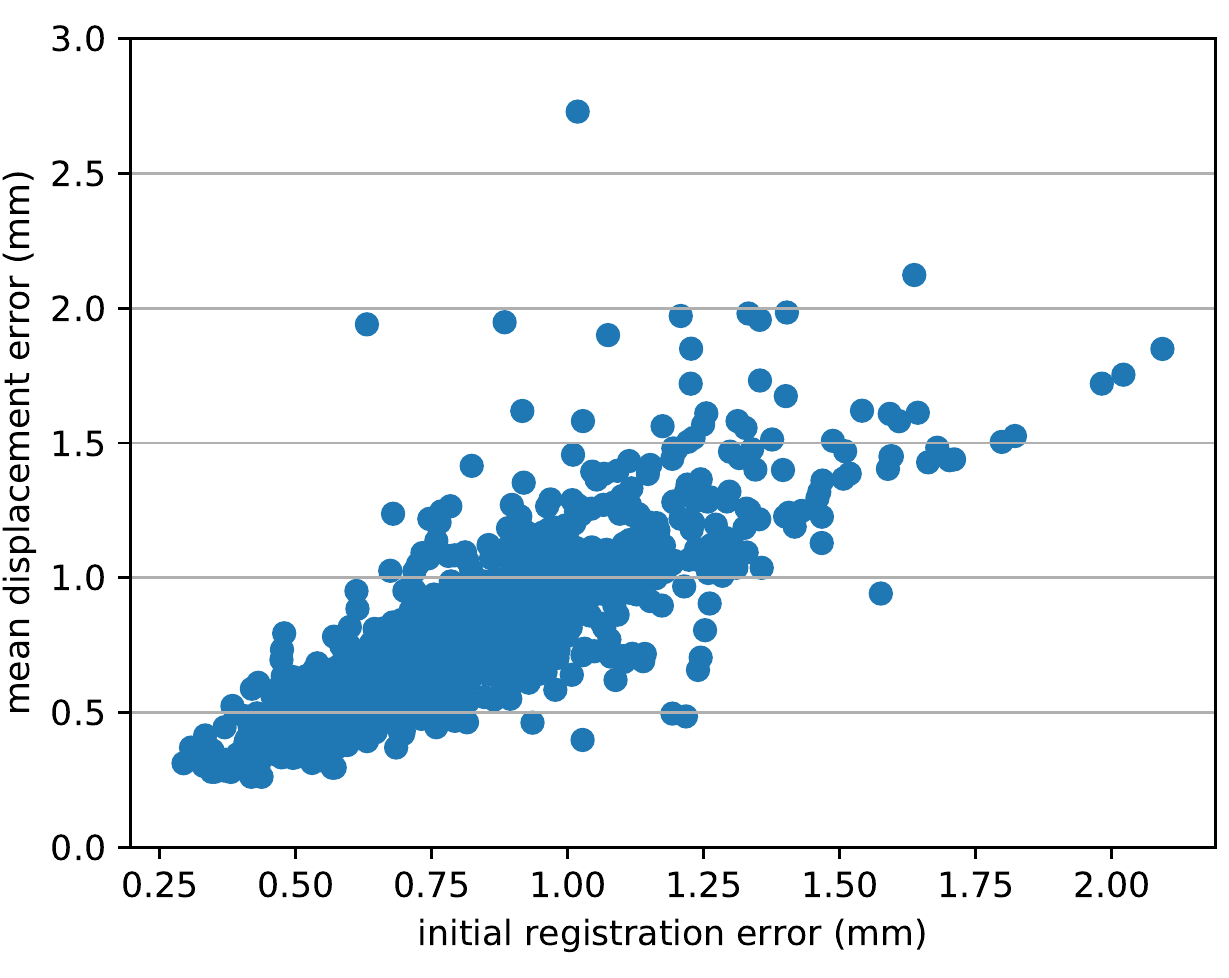}
    }

    \caption{a) shows the relation between the MDE of synthetic samples and the visible points ratio, which is calculated as the ratio between the point number of a target point cloud to the corresponding ground truth point cloud: $\lvert{P_{inv} }\lvert{ /} \lvert{\Tilde{T}}\lvert{} $. The decreasing trend shows that with more points available in the target in-vivo point clouds, the neural network tends to predict better deformation. b) depicts the MDE to the initial rigid registration error which is produced by the NgeNet. The better the initial rigid registration and correspondence set predicted by NgeNet, the lower the global non-rigid registration error from NDP becomes.} %
\label{evaluation:mde}
\end{figure}


\bmhead{In Silico}
A synthetic dataset is generated using the pipeline described in section \ref{method:syn_gen} with the same boundary conditions, and the target shape variants are unseen to the neural network during training. For each target shape, we estimated the displacement field for the identical template point cloud using C2P-Net trained on 20,000 samples. The mean target displacement field for the synthetic dataset is 1.54mm. Fig.~\ref{mde:visible_ratio} illustrates that the neural network tends to predict the deformation of the ex-vivo point cloud with higher accuracy if it obtains more information about the target in-vivo OCT point cloud. With few exceptions, Fig.~\ref{mde:initial_registration} shows when the initial registration error made by NgeNet is low, the final global non-rigid displacement field from NDP tends to be small.




\begin{figure}[!ht]
    \includegraphics[trim=0.1cm 2.7cm 5.3cm 3.3cm, clip,width=\textwidth]{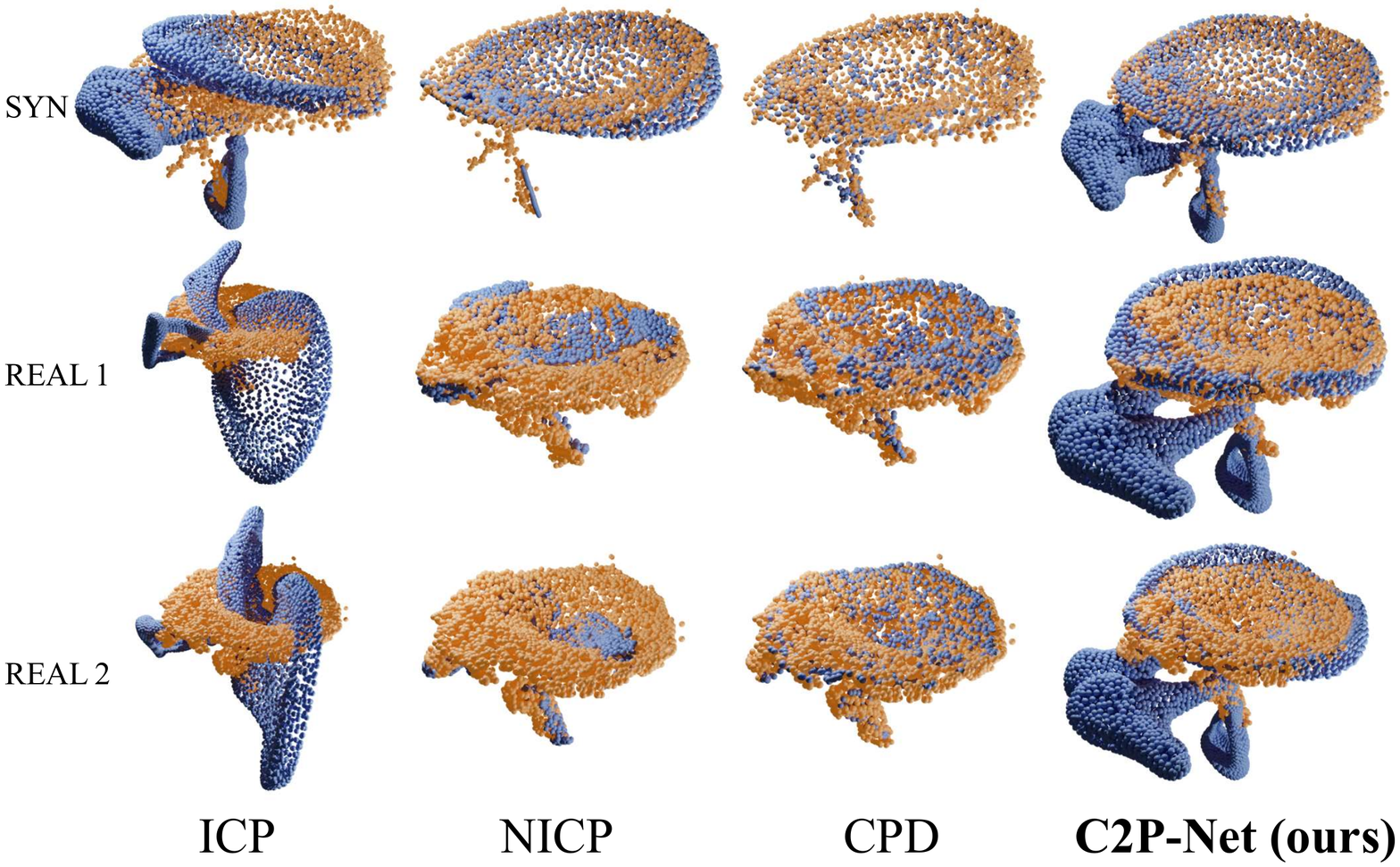} 
    \caption{Registration results of baseline models and C2P-Net on synthetic (SYN) and real OCT samples (REAL). Blue point clouds are the deformed ex-vivo ear models, i.e. the shape template, calculated by all the investigated methods. The orange point clouds stand for the registration target, i.e. in-vivo point clouds. From the visualization, we can see baseline non-rigid methods, NICP and CPD, tend to squeeze the points from the ex-vivo template largely to match the target point clouds, since they only focus on the position of local points, while C2P-Net retains the global geometry of the middle ear due to the learned correspondence knowledge between the ex- and in-vivo point clouds.} %
\label{evaluation:visualization}
\end{figure}

\bmhead{In Vivo}
A real OCT dataset with 9 samples of the human middle ear was collected with a handheld OCT imaging system\cite{Morgenstern2020} and segmented manually by surgeons within 3D Slicer along with a set of sparse landmarks for each structure of a sample (see Fig.~\ref{data_pipeline}). Since there is no ground truth correspondence available for the source and target point clouds to calculate $M_\mathrm{MDE}$, we introduce another metric based on the sparse landmarks:

\begin{equation}
    M_\mathrm{L}=\frac{1}{\lvert{L_{exv}}\rvert}\sum_{l_{i,k}\in{L_{exv}}}\operatornamewithlimits{min}_{l_{j,k}\in{L_{inv}}}\lVert{l_{i,k}-l_{j,k}}\rVert\:_{k=1,2,...,K}\\
    \label{landmarks_error}
\end{equation}
where $L_{exv}$ are the landmarks on the ex-vivo point cloud from $\mu$CT, which is also marked manually, $L_{inv}$ are the corresponding landmarks on the target in-vivo point cloud, and the landmark points belong to $K$ different segmentations.

Tab.~\ref{evaluation:stats_table} itemizes the registration results of C2P-Net and baseline models on both datasets. We can observe our registration pipeline outperforms the other methods in terms of $M_\mathrm{MDE}$ and $M_\mathrm{L}$. However, baseline models tend to have smaller $M_\mathrm{CD}$, which means the source point clouds are deformed largely to fit the target regardless of the anatomical geometry of the middle ear. Furthermore, this issue is demonstrated visually in Fig.~\ref{evaluation:visualization}, which depicts several registration results from C2P-Net and baseline models on real OCT samples. Hence, it is manifested that C2P-Net is able to register the partial target point cloud without losing the underlying geometry of source point clouds, which is the critical information clinicians want to obtain with registration. On the contrary, the baseline models only focus on the spatial position of local points regardless of the neighbouring and global geometry of the middle ear template.



\section{Conclusion}


In this work, we propose to improve the interpretation of OCT data for surgical diagnosis by fusing the knowledge from ex-vivo and general $\mu$CT data with the in-vivo noisy and incomplete OCT data. To this end, we propose C2P-Net which is based on NgeNet and Neural Deformation Pyramid. It takes the ex- and in-vivo 3D point clouds as input and explores the sparse correspondence between the two, then aligns them in a non-rigid fashion. Due to the lack of training data, a fast and effective synthetic simulation pipeline was designed using Blender3D, which produces noisy and partial point clouds as seen in-vivo based on the randomly generated shape variants of the middle ear. Our neural network was trained based on 20,000 synthetic samples, and it took around 3.5s to predict the displacement field for a given in-vivo point cloud at inference time.

To assess the performance of C2P-Net, experiments on synthetic and real OCT datasets were carried out. Since the real OCT data is noisy and incomplete, we investigated various metrics including mean displacement error, landmark error, and Chamfer distance and compared C2P-Net to the other baseline models from both statistics and geometry aspects. From Tab.~\ref{evaluation:stats_table} and Fig.~\ref{evaluation:visualization}, we can see that C2P-Net is able to deal with the partial OCT data and retain the anatomical structure during non-rigid registration, while other non-rigid methods do not understand the global geometry of the point clouds and focus only on local points. 


Future work will focus on improving registration accuracy. This can be achieved by improving the simulation pipeline with realistic noise to further bridge the gap between synthetic and real data. Furthermore, the segmentation information can be employed by the network to improve the prediction of correspondences. Apart from this, inference time can be further decreased by exploring the optimal configuration of the network for the OCT samples.

\backmatter





\bmhead{Acknowledgments}

Funded by the Else Kröner Fresenius Centre for Digital Health. 

\bibliography{reference}


\end{document}